\documentclass[runningheads]{llncs}

\usepackage{eccv}

\usepackage{eccvabbrv}

\usepackage{graphicx}
\usepackage{booktabs}

\usepackage[accsupp]{axessibility}  

\usepackage{cancel}

\usepackage{hyperref}

\usepackage{orcidlink}

\begin{document}

\title{OPPH: A Vision-Based Operator for Measuring Body Movements for Personal Healthcare} 

\titlerunning{A Vision-Based Operator}

\author{Longfei Chen \inst{1}\orcidlink{0000-0002-3935-8021} \and
Subramanian Ramamoorthy\inst{1}\orcidlink{0000-0002-6300-5103} \and
Robert B Fisher\inst{1}\orcidlink{0000-0001-6860-9371}}

\authorrunning{L Chen et al.}

\institute{The University of Edinburgh, 10 Crichton Street, Edinburgh, EH8 9AB, UK
\email{\{longfei.chen, s.ramamoorthy, r.b.fisher\}@ed.ac.uk}}

\maketitle

\begin{abstract}
Vision-based motion estimation methods show promise in accurately and unobtrusively estimating human body motion for healthcare purposes. However, these methods are not specifically designed for healthcare purposes and face challenges in real-world applications. Human pose estimation methods often lack the accuracy needed for detecting fine-grained, subtle body movements, while optical flow-based methods struggle with poor lighting conditions and unseen real-world data. These issues result in human body motion estimation errors, particularly during critical medical situations where the body is motionless, such as during unconsciousness.
To address these challenges and improve the accuracy of human body motion estimation for healthcare purposes, we propose the OPPH operator designed to enhance current vision-based motion estimation methods. 
This operator, which considers human body movement and noise properties, functions as a multi-stage filter. Results tested on two real-world and one synthetic human motion dataset demonstrate that the operator effectively removes real-world noise, significantly enhances the detection of motionless states, maintains the accuracy of estimating active body movements, and 
maintains long-term body movement trends. This method could be beneficial for analyzing both critical medical events and chronic medical conditions.
\end{abstract}

\section{Introduction}


Human movement measurement is widely used for healthcare purposes, such as gait analysis, rehabilitation and training, and chronic disease management \cite{hc1,hc2}. Unlike periodic clinical assessments, motion analysis systems ({e.g., wearables \cite{wear}, smartphone \cite{smartphone}, ambient motion sensors \cite{am1}, cameras \cite{miso}}) offer continuous, long-term monitoring of physical metrics in an objective manner. They help to obtain a quantitative assessment of motion parameters of the person \cite{wear}, enhancing reliability and optimizing resource allocation in healthcare settings.

Standard physical tests related to human motion include a range of motion tests, functional movement screening, the time up-and-go test, and the sit-to-stand test, etc. They highlight the importance of body movement measurements in various healthcare assessments. Automating these tests with accurate motion capture systems could significantly benefit clinicians and patients. Crisis events, such as loss of consciousness, paralysis, stroke, or heart attacks are frequent occurrences (globally) in daily life. Accurately identifying instances of these and providing timely intervention is crucial in these situations, particularly when individuals are alone.
Automated detection of these events could significantly enhance the chances of saving lives and alleviate the workload of caregivers. Furthermore, chronic conditions characterized by subtle changes in physical behavior over extended periods—such as age-related decline, frailty, dementia, arthritis, or post-surgical recovery—could greatly benefit from automatic motion analysis and monitoring systems. 

\begin{figure}[htbp]
    \centering
    \includegraphics[width=1\linewidth]{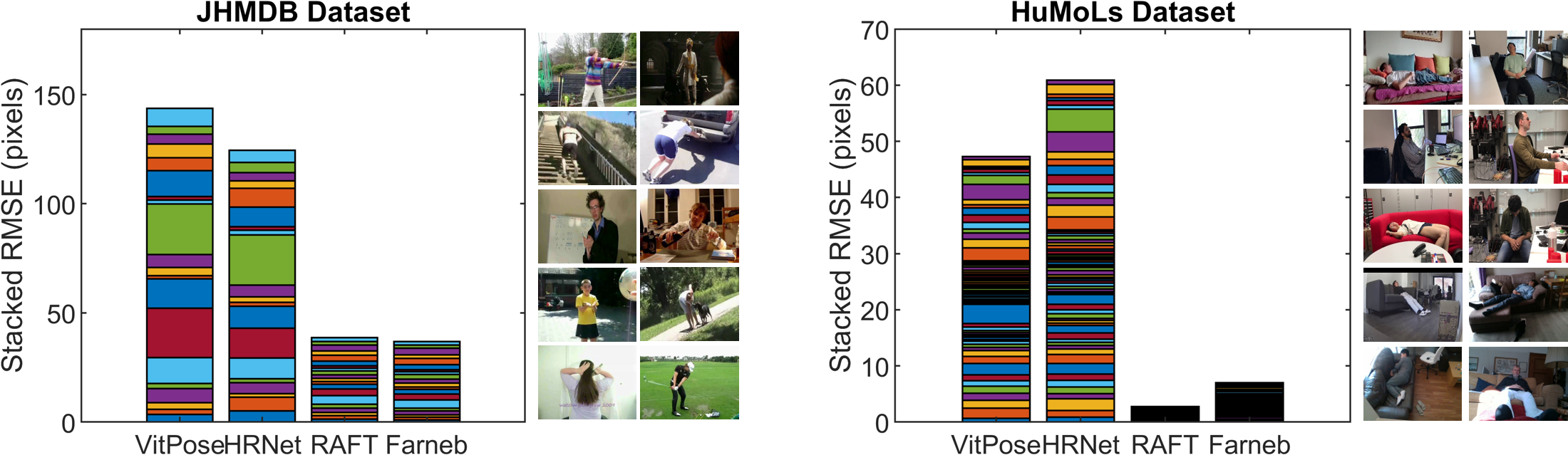}
    \caption{The estimation error of 2D body motion speed was evaluated for two real-world scenario datasets: one with only `active' body movement (JHMDB dataset \cite{jh}, containing 21 action classes), and another with `no' body movement (Human MotionLess dataset, HuMoLs, consisting of 67 videos from 18 subjects). The Root Mean Square Error (RMSE) is stacked across all classes (or videos). Four motion estimation methods (pose-based VitPose \cite{vit}, HRNet \cite{hr}, and optical flow-based RAFT \cite{raft}, Farneback \cite{farneback}) were compared to the ground truth. Pose-based methods showed larger errors in both motion and motionless scenarios compared to optical flow-based methods. 
}
    \label{fig:orierror}
\end{figure}

Computer vision-based motion analysis approaches are potentially beneficial, such as human pose estimation and optical flow.
Compared to wearable systems, these methods 
require no direct interaction, while providing clinically acceptable accuracy \cite{clinical,clinical2}, and are suitable and reliable for long-term measurements. Pose-based methods estimate kinematic information like the locations of human joints, which provide high-level features, e.g., body parts, and are beneficial for action recognition \cite{p8}, but may not be fine-grained enough to measure subtle movements. On the other hand, optical flow-based methods can estimate a dense motion field, which can provide subpixel-level accuracy of the motion. 

However, both pose-based and optical flow methods face challenges related to body movements in real-world healthcare applications. Current vision-based methods are not specifically designed for healthcare purposes and cannot meet some requirements, such as accurately detecting motionless events and long-term subtle changes. Specifically, the challenges include:

(1) Obtaining good motion ground truth for training. Most existing datasets related to human body optical flow are synthetic, like MPI Sintel \cite{sintel}, and do not include real-world artifacts, including image sensor errors, illumination variations, poor lighting, and image blur. Manually creating ground truth, such as labeling each frame with body models \cite{jh} or joint locations \cite{mpii}, may also include annotation inaccuracies. 
(2) Lack of adaptability to real-world scenarios \cite{TM}, such as unseen environments or rare human poses. Fine-tuning these models for new real-world scenarios is considered challenging, as gathering ground truth training data for every new testing scenario is difficult. 
(3) Most previous works and existing vision-based datasets \cite{jh,hm51,sintel,h36m_pami} focus on general human body movements, evaluating the accuracy of measuring `active motion' or `actions' of humans, while `inactive motion' or `motionless' situations are mostly ignored. However, motionless situations are important for indicating crisis medical events, such as loss of consciousness, heart attack, or stroke.
\Cref{fig:orierror} shows the estimation accuracy of 2D body motion speed using two pose-based methods \cite{vit,hr} and two optical flow-based methods \cite{raft,farneback} on two real-world datasets. It shows that pose methods generally exhibit larger errors when measuring 2D body part speeds compared to optical flow-based methods. 
However, both pose and optical flow-based methods have limitations in accurately estimating motionless body events. These errors can lead to difficulties in distinguishing subtle body movements from noise in real-world applications, and may also result in missing critical medical events related to motionless body detection and inaccuracies when measuring long-term motor changes.

This study proposes the computer vision operator OPPH, specifically designed for human body motion estimation for healthcare purposes. 
The aim is to enhance vision-based motion estimation methods and tailor them for healthcare applications, making them effective for both crisis medical situations and long-term chronic health monitoring.
The operator aims to (1) accurately estimate active body movement, (2) accurately detect body motionlessness, and (3) accurately capture the long-term human body movement trends.
We compare the operator with state-of-the-art (SOTA) pose-based \cite{vit,hr} and optical flow-based \cite{raft,farneback} motion estimation methods, and test for accuracy in estimating 2D human body motion speed. The results show that:\\
(a) For our collected real-world human motionless dataset (HuMoLs), the OPPH operator almost completely removed the error for pose-based and optical flow-based motion estimation methods, achieving the best mean RMSE of 
$2.7 \times 10^{-4}$ pixels for 2D body motion speed estimation, surpassing other commonly used denoising filters by a large margin.\\
(b) For a small-scale real-world human action dataset \cite{jh} and a large-scale synthetic human motion dataset \cite{su}, the OPPH operator enhanced the accuracy of pose-based methods, while achieving comparable performance to that of optical flow-based methods.\\
(c) The OPPH operator, along with the SOTA RAFT optical flow-based method \cite{raft}, can effectively 
{maintain} long-term trends and changes in body movements (correlation R = 0.92-0.99).

\section{Related Works}
Various human body motion estimation methods are employed in healthcare. 
Wearable sensors \cite{wear,wear2,wear3,wear4}, such as IMUs, can give highly accurate human motion estimation, typically with joint position errors ranging from 4 to 15 mm and angle estimation errors of 1 to 5 degrees \cite{wear,wear4}. However, wearing multiple sensors can be intrusive, and frequent recharging may be required \cite{wear3}. Additionally, sensor placement can introduce noise, drifting, and soft tissue artifacts \cite{wear}. Ambient sensors \cite{am1,am2} in smart environments, such as pressure sensors on floors, infrared and motion sensors, radar, and ultrasonic sensors, provide unobtrusive monitoring of human movement over broader areas and effectively detect gross movements like walking or standing. Ambient sensors capture less detailed information about specific body parts or precise movements, compared to wearables, marker-based methods \cite{mk11}, or camera-based systems \cite{cam1}. 
Marker-based motion capture systems provide high accuracy, with position errors typically below 1 mm and angular errors of about 1 degree \cite{mk1,mk2}.
Marker-less motion capture systems, which do not require physical contact, are widely used for analyzing human body movement in real-world clinical settings \cite{mkless1,mkless2,mkless3,mkless4}. These systems utilize computer vision techniques to track body movements, offering clinically acceptable accuracy typically within positional errors of 10 to 50 mm and angle errors of 2 to 5 degrees \cite{clinical,clinical2,cvacc1,cvacc2}.
For example, pose-based kinematic estimation methods \cite{vit,hr} are commonly utilized for remote tracking of body joints during physiotherapy rehabilitation exercises \cite{p1}, gait analysis \cite{p2,p6}, telehealth applications \cite{p3}, assessing child motor skill development \cite{p4}, lower body rehabilitation \cite{p5}, and evaluating neurological diseases \cite{p6,p7}.
Pose-based methods, which rely on detecting and tracking predefined key points and joints, may be less sensitive to subtle body movements such as finger-level motions.

Several optical flow methods can be used for human body motion estimation. Traditional methods include Lucas-Kanade \cite{lucas} and Farneback \cite{farneback}, while deep learning-based approaches such as FlowNet \cite{flownet}, RAFT \cite{raft}, and PWC-Net \cite{pwc} are also widely used. These methods are generally designed for broader applications rather than specifically optimized for estimating human body motion. Few studies have applied optical flow to healthcare-related purposes. For instance, Kashevnik et al. \cite{op1} used optical flow to estimate human body motion and respiratory characteristics.

Several datasets are available for human body motion analysis, providing ground truth data for both pose and optical flow. Examples include the JHMDB dataset \cite{jh}, which encompasses 21 actions in real-world scenarios. Additionally, the Surreal dataset \cite{su} and the Human Flow dataset \cite{lh} feature synthetic human body models against realistic backgrounds, offering ground truth motion data.

Various methods have been used to enhance the accuracy of motion estimation further. The most common methods include pre/post-processing spatial and temporal smoothing, such as Gaussian filtering \cite{opfil}, Bilateral filtering \cite{bif}, and the use of regularization techniques such as smoothness \cite{hs} or Total Variation regularization \cite{TV2}. In this study, we proposed an operator to improve motion estimation methods and compared it with these filters.

\section{Methodology}

The proposed OPPH operator aims to improve the estimation of human body motion for healthcare applications. 
It assumes that information about the body region (body mask) and body movement (such as optical flow or pose) has already been given \textcolor{blue}{\cite{jh, su}}, or obtained through other techniques, \textcolor{blue}{e.g., \cite{vit,hr} or \cite{raft,farneback}.}

\begin{figure}[htbp]
    \centering
    \includegraphics[width=1\linewidth]{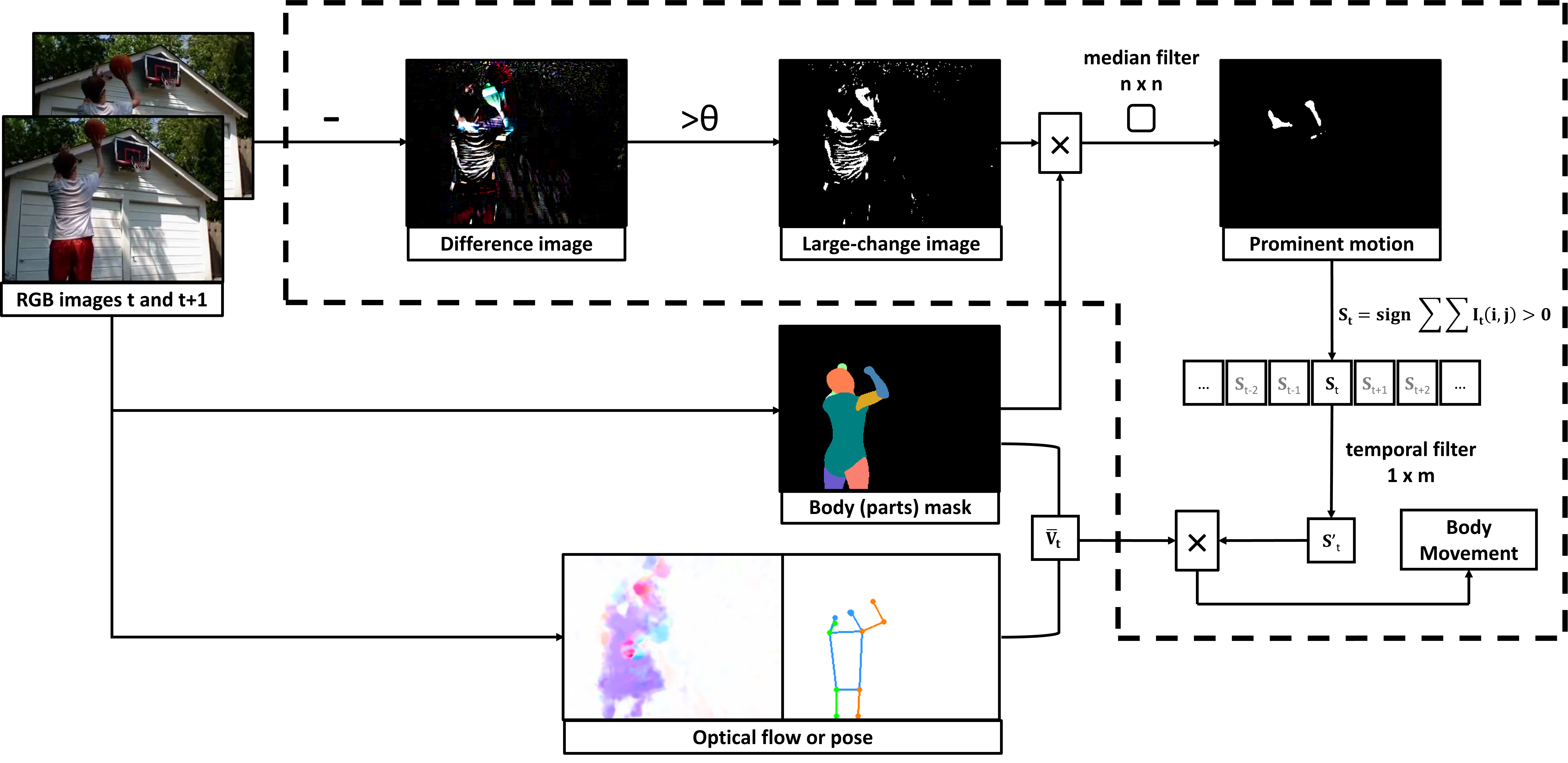}
    \caption{
    {The proposed OPPH operator for enhancing human body motion estimation consists of thresholding $\theta$, an $n \times n$ spatial filter, and a $1 \times m$ temporal filter. First, three-color-channel thresholding is applied to the difference image derived from two consecutive RGB images to identify large-change pixels. The large-change image is then masked by a pre-obtained human body mask and processed through a spatial filter to remove isolated noise. Next, the prominent motion image is compressed into a single binary variable $S_t$, which is temporally filtered to produce $S_t'$, indicating whether frame $t$ contains true body movement. Finally, the filtered signal $S_t'$  is used to gate the original body motion speed estimation (derived from optical flow magnitudes or pose displacements and the body mask) to obtain the final body motion speed estimation result for each frame.}
    }
    \label{fig:theoperator}
\end{figure}

As shown in \cref{fig:theoperator}, the OPPH operator has several steps:\\
1. OPPH takes two consecutive RGB image frames and calculates the difference image by subtracting corresponding pixels between the frames.\\
2. A threshold ($\theta$) is applied to color channels (red, green, and blue) of the difference image. If the absolute value of the difference in all color channels is greater than the threshold, the corresponding pixel in the output image $I(i,j)$ is set to 1. Otherwise, it's set to 0. This step essentially identifies large changes in pixel intensity between the frames, based on the assumption that real body movement involves changes in the brightness and color intensity of the image. This threshold, learned from real-world videos, is chosen to distinguish subtle body movements from environmental noise under various lighting conditions (details can be found in \cite{miso}).\\
3. The large-change image is multiplied by the pre-computed body mask. An $n \times n$ median filter is applied to remove isolated noisy pixels from the masked large-change image. This step aims to eliminate noise caused by camera sensors, poor lighting conditions, or background movement (like leaves rustling in the wind, see \cref{fig:theoperator}). The remaining pixels are considered to represent prominent motion regions. This is based on the assumption that real body movement is a connected region (blob) rather than isolated pixels.
The result of this step is a binary image $P_t$ indicating the presence of prominent motion for each frame $t$.
 \\
{4. We now compress the prominent motion image $P_t$ to a single binary variable $S_t$. The binary pixel values in $P_t$ are added together, and if the sum is non-zero, then $S_t$ is set to 1, and otherwise it is set to 0.
A temporal median filter $1 \times m$ is then applied to $S$ to give $S'$. This performs a temporal filtering to remove temporary flickering. 
If the value $S_t'$ of the filtered prominent motion has value 1, 
the frame is considered to have observed true human body movement.
Otherwise, it's considered noise. 
This step helps to distinguish short-lived noise from actual body movement. 
This is based on the assumption that true human body movement lasts longer than environmental noises such as sudden changes in illumination (e.g., flickering TV light reflecting on the body).}\\
{5. The optical flow field between frames is calculated to obtain the speed of movement. Specifically, the magnitudes of the optical flow vectors are multiplied by the pre-derived body mask to focus only on body movements, ignoring the background, and then the result is averaged to a single value to determine the overall speed of movement for the frame. For pose estimation data, the displacements in all pose joints are similarly averaged to a single value representing body movement speed for the frame. This resulting speed value is then multiplied by $S_t'$ to get the final speed result for the frame. If $S_t'$ is 1, the frame is considered to contain true body movement, and the calculated speed is recorded; if $S_t'$ is 0, the frame is considered noise and the speed is set to zero.}

\section{Experimental Results}

Three experiments below will show that:
\begin{enumerate}
    \item The OPPH operator improved the accuracy of the pose-based methods and gave a comparable performance on the optical flow-based methods  (\cref{expt12}).
    \item OPPH almost completely removed motion errors in the HuMoLs dataset (\cref{expt12}).    \item {OPPH maintained the long-term trends and changes in body movements estimated using a state-of-the-art method.} (\cref{expt3}). 
\end{enumerate}

Three datasets are used to evaluate the effectiveness of OPPH in enhancing the existing human body motion estimates:\\
\textbf{JHMDB} \cite{jh}: A real-world dataset for 21 human actions. It includes indoor active actions such as brushing hair, drinking, and clapping plus outdoor actions such as jumping, walking, and kicking a ball. 
It provides human 2D poses, optical flow, and ground truth for human body parts. Poses are manually labeled using a 2D puppet model, and human optical flow is labeled using puppet flow \cite{ppflow}. A total of 888 videos are contained in this dataset.\\
\textbf{HuMoLs}: A real-world human \emph{motionless} dataset collected locally, comprising 67 videos from 18 subjects in different indoor locations worldwide, captured using various devices and imaging resolutions. 
{In all videos, the observed person is motionless.}
The ground truth for 2D human body motion speed is set to zero.
{The HuMoLs dataset {can be found at \url{https://groups.inf.ed.ac.uk/vision/DATASETS/HUMOLS/}.}\\
\textbf{Surreal} \cite{su}: A large-scale synthetic person dataset with realistic backgrounds. It provides ground truth for depth, body parts, optical flow, and pose. It includes videos with large body movements, tiny body movements, and motionless bodies. A total of 5700 videos of the dataset are tested in this study.

\textbf{Parameters}: The optimal intensity difference threshold $\theta$ is learned as 20 to 25 for all RGB channels (0-255) \cite{miso}. Here, 20 is chosen for the experiments. The spatial filter size $n$ is set to 5 for image size over 640$\times$480 for Surreal and HuMoLs; and set to 3 for image size 320$\times$240 for JHMDB. And the temporal window $m$ is set as $fps/2$, i.e., it keeps motion longer than 0.5 seconds.

\textbf{Evaluation metric}: The mean 2D human body motion speed in each frame \( t \) is represented as:

\[
\bar{V}_t = \frac{1}{N} \sum_{(i,j) \in \text{BodyMask}} \sqrt{V_x(i,j)^2 + V_y(i,j)^2}
\]
where \((i,j)\) indexes over all pixels in the body mask, \((V_x(i,j), V_y(i,j))\) is the optical flow vector at each pixel \((i,j)\), and \(N\) is the number of pixels within the body mask.

For the pose-based method, the mean 2D body motion speed is calculated similarly, but \((V_x, V_y)\) are the displacements in joint locations between two frames, and \(N\) is the number of body joints:

\[
\bar{V}_t = \frac{1}{N} \sum_{k=1}^{N} \sqrt{V_x(k)^2 + V_y(k)^2}
\]

The error rate is calculated using the RMSE between the estimated 2D body motion speed and the ground truth speed:

\[
\text{RMSE} = \sqrt{\frac{1}{n} \sum_{t=1}^{n} ( \bar{V}_{t,\text{EST}} - \bar{V}_{t,\text{GT}} )^2}
\]

\noindent 
where \(\bar{V}_{t,\text{EST}}\) is the estimated mean 2D body motion speed in frame \( t \), \(\bar{V}_{t,\text{GT}}\) is the ground truth mean 2D body motion speed at frame \( t \), and \(n\) is the total number of frames in the video clip. The mean RMSE of a dataset is the average RMSE of its videos (or classes).

Two pose estimation methods, VitPose \cite{vit} and HRNet \cite{hr}, and two optical flow-based methods, RAFT \cite{raft} and Farneback \cite{farneback}, are tested with and without the proposed OPPH operator. 
The ground truth 2D body motion speed for both pose-based and optical flow-based methods is derived from the labeled ground truth data of the datasets.
{The average speed is also calculated for each body part, where the ground truth masks are provided in the JHMDB and Surreal datasets.}

The 2D motion speed was chosen as the unified criterion to compare different types of methods, rather than using their specific measurement metrics such as Endpoint Error (EPE) for optical flow or Percentage of Correct Keypoints (PCK) for pose-based methods.
The {average} motion speed of the body (and body parts) is also more meaningful for healthcare applications.

\begin{figure} [htbp]
    \centering
    \includegraphics[width=1\linewidth]{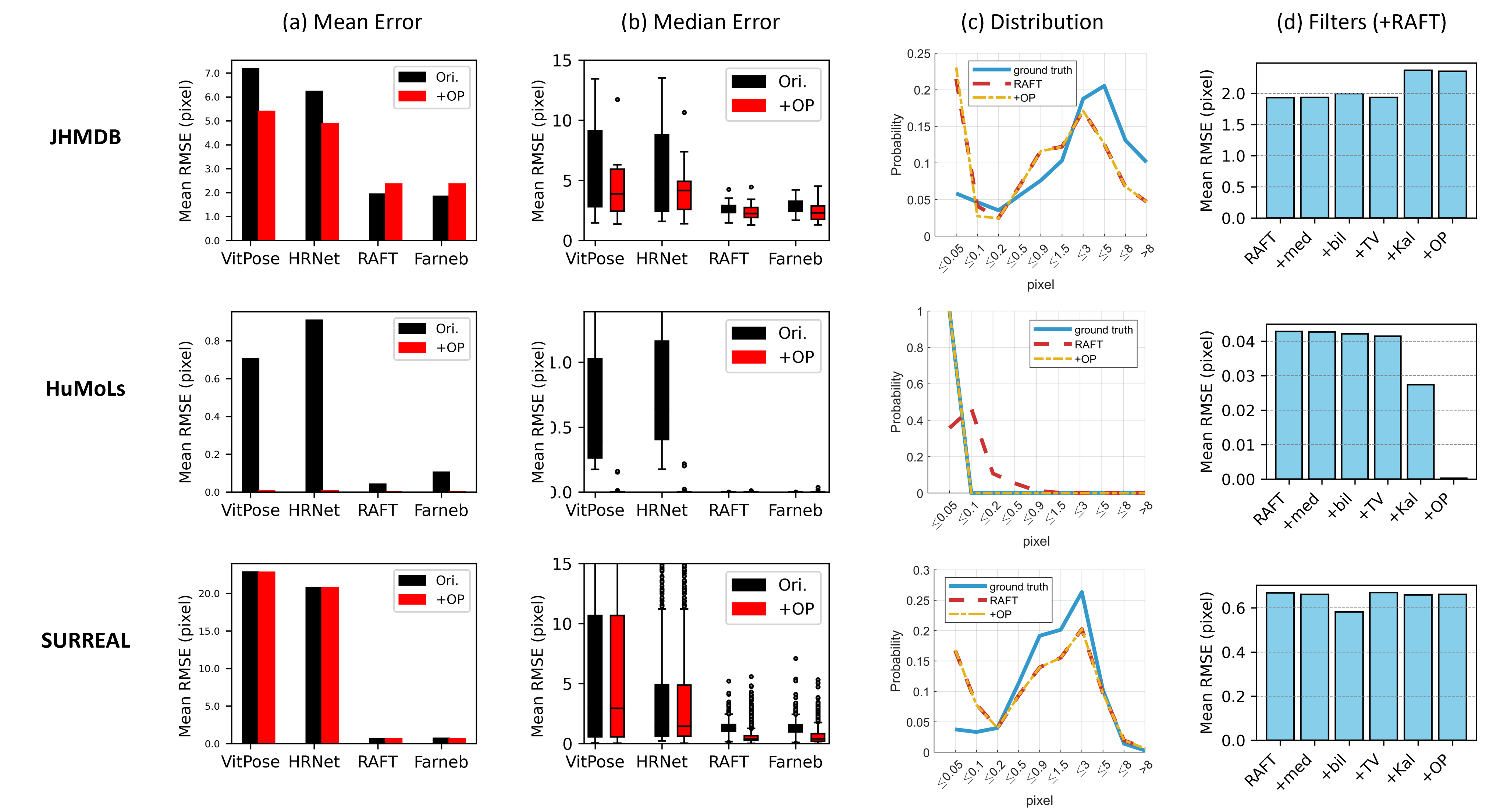}
    \caption{
    Accuracy of 2D human body motion speed estimation. (a) Motion speed estimation error (mean RMSE across classes/videos) with all datasets and four original motion estimation methods (Ori) and with OPPH (+OP). (b) The median RMSE (outliers excluded) for the same comparison as in (a). (c) Comparison of speed distributions between the ground truth motion speed and the estimated speed by the best method, RAFT, and RAFT with OPPH (+OP). (d) Accuracy (mean RMSE) of OPPH compared with other types of filters on RAFT. OPPH (+OP) was compared with the Median filter (+med), Bilateral filter (+bil), Total Variation (+TV), and Kalman Filter (+kal).
    }
    \label{fig:accuracies}
\end{figure}

\subsection{Motion estimation accuracy\label{expt12}}
\textbf{Estimation error}: \Cref{fig:accuracies}(a) shows the mean RMSE for 2D body motion speed estimation. On the JHMDB dataset, OPPH significantly improved the accuracy of the SOTA pose-based methods, reducing the mean RMSE by 1.78 pixels for VitPose and 1.34 pixels for HRNet. For optical flow-based methods, OPPH slightly increased the mean RMSE by 0.42 pixels for RAFT and 0.51 pixels for Farneback. 
\Cref{fig:accuracies}(b) demonstrates that OPPH enhances the typical performance of both the pose-based methods and the optical flow-based methods by reducing the median RMSE (outlier removed). Note that there will be inaccuracies in the manually annotated ground truth for the real-world datasets.

On the real-world human motionless dataset HuMoLs, the OPPH operator significantly reduced the error rate for 2D body motionless estimation across all SOTA methods. The mean RMSEs were $5.1 \times 10^{-3}$ pixels for VitPose, $6.9 \times 10^{-3}$ pixels for HRNet, $2.7 \times 10^{-4}$ pixels for RAFT, and $8.6 \times 10^{-4}$ pixels for Farneback. The remaining small errors in the HuMoLs dataset are attributed to inaccuracies in the ground truth labeling. Although the ground truth labels indicate complete motionlessness, some slight body movements, such as eye blinking or slight body adjustments by the subjects, were present during data capture.

For the Surreal dataset, where real-world noise is absent, the OPPH operator produced only minor improvements for all methods. 
The mean RMSE decreased by 0.035 pixels for pose-based methods and by 0.025 pixels for optical flow-based methods.

The numerical accuracy of all methods on all datasets, with or without OPPH, is presented in \cref{tab:acc}.
Overall, optical flow-based methods had significantly lower error rates for 2D human body motion speed estimation compared to pose-based methods.
Across two real-world datasets and one synthetic dataset, the optical flow-based RAFT method, when combined with OPPH, achieved the best overall performance.

\textbf{Distribution}: \Cref{fig:accuracies}(c) illustrates the distribution of 2D body motion speeds for the ground truth and RAFT (with and without OPPH). Compared to the ground truth, OPPH effectively suppressed estimation errors for smaller motion speeds, {i.e., suppressed movement speeds between 0.1 to 0.2 pixels/frame for JHMDB, and in the range of 0 to 0.9 pixels/frame for HuMoLs.}


\textbf{Comparison with filters}: The proposed OPPH operator was compared with several commonly used post-processing denoising filters applied to the motion field of RAFT, including Median filter, Bilateral filter, Total Variation, and Kalman Filter.
\Cref{fig:accuracies}(d) shows that, on the JHMDB dataset, similar to the mean RMSE in (a), OPPH  slightly decreased the original performance of RAFT compared to other filters. On the HuMoLs dataset, only the Kalman Filter and OPPH effectively reduced the error, with OPPH surpassing all filters by a large margin, nearly eliminating all errors. On the Surreal dataset, only the Bilateral filter showed a slight decrease in error, while OPPH and other filters did not have significant benefits on this synthetic dataset.

\subsection{Correlation analysis\label{expt3}}

\Cref{fig:correlations} depicts the long-term trend analysis of body movement using RAFT and the OPPH operator. For the JHMDB dataset, the average 2D motion trend was calculated every 1, 2, and 3 minutes due to the small dataset size and short video lengths. For the longer videos in the Surreal dataset, the trend was calculated every 1, 5, 10, 20, 30, and 60 minutes.
{The Pearson correlation coefficient $r$ is calculated between the estimated value and the ground truth value, for the motion speed in each time window.}
The results, as summarized in \cref{tab:corr}, show that OPPH largely maintains the original correlation.
We hypothesize that the slight degradation on the SURREAL dataset is due to the presence of small ground truth motions that are suppressed, whereas there is no real-world noise to suppress.



\begin{figure}
    \centering
    \includegraphics[width=.8\linewidth]{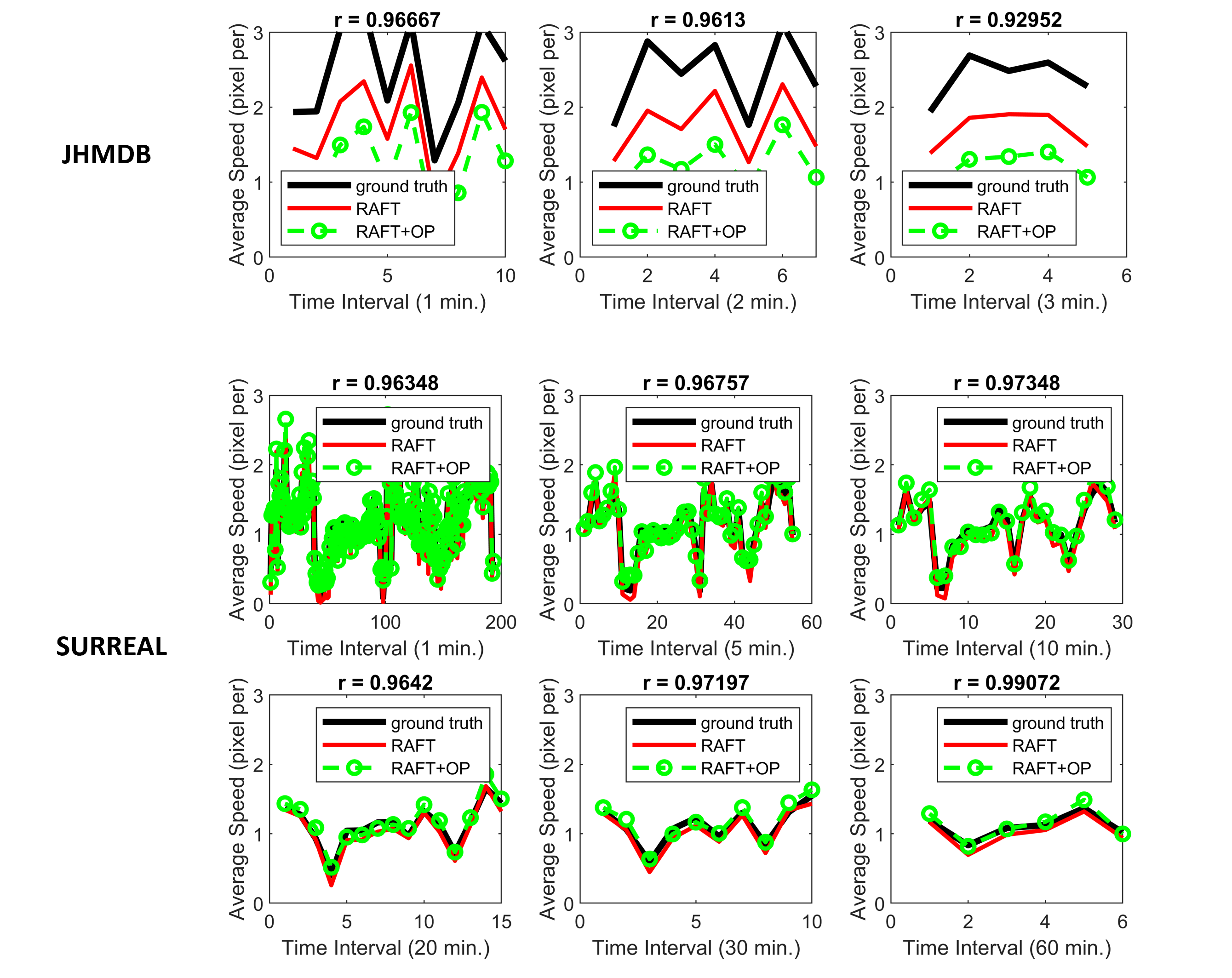}
    \caption{Long-term correlation between the estimated 2D body motion and the ground truth motion. The videos in the two motion datasets are stacked into longer time windows, and the mean motion speed within each window is calculated (pixels per time window). 
    Each plot averages the data for a different period, as reported under the plot. The horizontal axis is the index of the period.
    The results show that OPPH (+OP) combined with the optical flow-based method, RAFT, can still effectively capture the trend of body movement changes.
    {From the distribution in \cref{fig:accuracies}, there is an underestimation of fast movement in JHMDB by RAFT, and the OPPH operator further suppressed the RAFT estimated values in low-speed ranges.
    This explains the offset between the curves for JHMDB, however note that there is still a high correlation. For Surreal, RAFT underestimated the middle-range speed compared to the ground truth but overestimated the low-range speed; OPPH does not have any substantial effect on the estimated value.}
    }
    \label{fig:correlations}
\end{figure}

In conclusion, the proposed OPPH operator effectively mitigated real-world noise, enhancing the accuracy of both SOTA pose-based and optical flow-based human body movement estimation methods. This improvement was particularly notable during periods of human motionlessness. Furthermore, OPPH, in conjunction with motion estimation methods, 
{maintains}
the long-term trends of the human body movements, reinforcing its utility for movement analysis.

\subsection{Processing speed\label{expt4}}
The operator's real-time performance was evaluated on a desktop PC running Python 3.8, with an 8-core CPU at 2.5 GHz and an RTX 3080 GPU. The processing times were as follows: at a resolution of 640x480, human mask inference (YoloV8n-seg \cite{yolov8}) took approximately 30 ms, dense optical flow calculation (Farneback \cite{farneback}) took around 30 ms, and the OPPH operator added an additional 15 ms. This setup achieves a real-time performance of approximately 13.33 frames per second.


\begin{table}[]
\centering
\caption{The numerical accuracy of the four methods for 2D body motion speed estimation, without and with the OPPH operator (+OP).This includes Mean RMSE (top) and Median RMSE (bottom). The bold numbers show the best performance for each dataset.}

\begin{tabular}{ccccc}
\hline
Mean   RMSE  & VitPose\cite{vit} & HRNet\cite{hr}  & RAFT\cite{raft}  & Farneback\cite{farneback} \\
\hline
\hline
JHMDB\cite{jh}       & 7.181   & 6.224  & 1.932 & \textbf{1.842}     \\
JHMDB\cite{jh}+OP       & 5.396   & 4.883  & 2.354 & 2.360     \\
\hline
HuMoLs      & 0.706   & 0.909  & 0.041 & 0.104     \\
HuMoLs+OP      & 0.005   & 0.007  & \textbf{$\boldsymbol{2.7} \boldsymbol{\times} \boldsymbol{10^{-4}}$} & $8.6 \times 10^{-4}$     \\
\hline
SURREAL\cite{su}      & 22.857  & 20.796 & 0.668 & 0.701     \\
SURREAL\cite{su}+OP     & 22.824  & 20.760 & 0.661 & \textbf{0.658}     \\

\hline
           \hline
\end{tabular}
\ \\
\ \\
\ \\
\begin{tabular}{ccccc}
\hline
Median RMSE        & VitPose\cite{vit} & HRNet\cite{hr}  & RAFT\cite{raft}  & Farneback\cite{farneback} \\
\hline
\hline
JHMDB\cite{jh}       & 5.224   & 4.978  & 2.871 & 2.510     \\
JHMDB\cite{jh}+OP       & 3.882   & 4.162  & 2.305 & \textbf{2.242}     \\
\hline
HuMoLs      & 0.394   & 0.640  & 0.000 & 0.000     \\
HuMoLs+OP      & \textbf{0.000}   & \textbf{0.000}  & \textbf{0.000} & \textbf{0.000}     \\
\hline
SURREAL\cite{su}      & 2.955   & 2.955  & 0.526 & 0.731     \\
SURREAL\cite{su}+OP     & 2.932   & 2.932  & 0.403 & \textbf{0.323}    \\
\hline
\hline
\end{tabular}
\label{tab:acc}
\end{table}

\begin{table}[]
\centering
\caption{The correlation {between the ground truth and RAFT, with and without the OPPH post-processing, and on both the JHMDB and SURREAL datasets.}
}
\begin{tabular}{ccccccc}
\hline
JHMDB\cite{jh}   & 1 minute  & 2 minutes  & 3 minutes  &        &        &        \\
\hline
\hline
RAFT\cite{raft}    & 0.9694 & 0.9639 & 0.9092 &        &        &        \\
RAFT\cite{raft}+OP & 0.9667 & 0.9613 & 0.9295 &        &        &        \\
\hline
\hline
&&&&&&\\
\hline
SURREAL\cite{su} & 1 minute  & 5 minutes  & 10 minutes & 20 minutes & 30 minutes & 60 minutes \\
\hline
\hline
RAFT\cite{raft}    & 0.9754 & 0.9842 & 0.9883 & 0.9895 & 0.9849 & 0.9947 \\
RAFT\cite{raft}+OP & 0.9635 & 0.9676 & 0.9735 & 0.9642 & 0.972  & 0.9907\\
\hline
\hline
\end{tabular}
\label{tab:corr}
\end{table}

\section{Conclusion and Discussion}

In this study, we introduced the OPPH operator designed to improve human body motion estimates, which would be beneficial for healthcare applications. 
Acting as a multi-stage filter, OPPH reduces real-world noise, thereby clarifying true body movements. 
It can be integrated into both deep learning-based structures and traditional energy-based optical flow methods (as a weighted regularization term).

The findings demonstrate that OPPH effectively reduces noise associated with human motionlessness in state-of-the-art motion estimation methods while preserving the accuracy of 2D body motion speed estimates for active body parts. 
In healthcare contexts, OPPH's ability to minimize motionless noise is particularly valuable, as it can be critical for identifying certain health crises. Furthermore, OPPH maintains the long-term trend of body movements, making it useful for purposes such as chronic condition monitoring. 
In future work, this technique could be applied to practical scenarios for monitoring motionlessness, inactivity, and deterioration in real-world situations, such as supervising children in kindergarten, infants riding in cars or buses on hot days, elderly individuals staying home alone, and patients undergoing treatment or recovery.

\section*{Acknowledgement}
This research was funded by the Legal \& General Group (research grant to establish the independent Advanced Care Research Centre at the University of Edinburgh). The funder had no role in the conduct of the study, interpretation or the decision to submit for publication. The views expressed are those of the authors and not necessarily those of Legal \& General. 

%
%
\bibliographystyle{splncs04}
\bibliography{main}
\end{document}